\documentclass[a4paper,fleqn]{cas-sc}

\usepackage{graphicx}
\usepackage{algorithm}
\usepackage{algpseudocode}
\usepackage{tcolorbox}
\tcbuselibrary{breakable}
\usepackage{amsmath}
\usepackage{subcaption}
\usepackage{multirow}
\usepackage{booktabs}
\usepackage{makecell}
\usepackage{threeparttable}

\usepackage{nomencl}
\makenomenclature
\usepackage{etoolbox}
\renewcommand\nomgroup[1]{%
  \item[\bfseries
  \ifstrequal{#1}{A}{Sets and indices}{%
  \ifstrequal{#1}{B}{Parameters}{%
  \ifstrequal{#1}{C}{Binary variables}{}}}%
]}
\usepackage[authoryear]{natbib}

\begin{document}
\let\WriteBookmarks\relax
\def\floatpagepagefraction{1}
\def\textpagefraction{.001}

\shorttitle{Efficient Detailed Scheduling of Multiproduct Pipelines using Action-Decoupled DRL}
\shortauthors{K. Shao et~al.}

\title[mode=title]{Efficient Detailed Scheduling of Multiproduct Pipelines using Action-Decoupled Deep Reinforcement Learning}

\author[1]{Kai Shao}
\ead{Z19090002@s.upc.edu.cn}
\ead[orcid]{0009-0001-3687-3108}

\author[1]{Ling Jian}
\ead[orcid]{0000-0002-9385-5977}
\ead{bebetter@upc.edu.cn}

\author[1]{Liming Liu}
\ead{s25080030@s.upc.edu.cn}

\author[2]{Serguey V. Bogomolov}
\ead[orcid]{0000-0002-5414-6019}
\ead{bogomo@cs.msu.su}

\author[1]{Xinmin Wang}
\cormark[1]
\ead{wangxinmin@upc.edu.cn}
\ead[orcid]{0000-0002-4050-0865}

\affiliation[1]{organization={School of Economics and Management, China University of Petroleum (East China)},        city={Qingdao},
                state={Shandong},
                country={China}}

\affiliation[2]{organization={Faculty of Computational Mathematics and Cybernetics, Lomonosov Moscow State University},
                addressline={Leninskie Gory},
                city={Moscow},
                postcode={119899},
                country={Russia}}

\cortext[1]{Corresponding author}

\begin{abstract}
Multiproduct pipelines transport refined products to delivery stations in sequential batches, and their detailed scheduling must meet market demand while respecting nonlinear hydraulic constraints. Retaining the full hydraulic nonlinearity yields a mixed-integer nonlinear program that is computationally intractable, while the linearization commonly adopted for solver-based optimization degrades hydraulic accuracy without resolving the heavy computational burden. This paper develops an action-decoupled proximal policy optimization (PPO) framework that efficiently produces detailed schedules while preserving hydraulic fidelity. The scheduling process is cast as a semi-Markov decision process (SMDP) with flexible action durations. To handle the discrete-continuous hybrid action space, a cascaded dual-actor architecture decouples the discrete operating duration from the continuous flow decisions, eliminating the gradient interference that destabilizes conventional coupled-action policies and thereby yielding high-quality schedules far more efficiently. Validated on a real-world multiproduct pipeline in China and benchmarked against a single-actor baseline and a linearized MILP solved by Gurobi, the proposed framework achieves delivery accuracy close to the exact optimum while substantially improving flow rate stability and substantially reducing solution time.
\end{abstract}


\maketitle


\section{Introduction}\label{sec1}
\subsection{Background}
The petroleum supply chain encompasses multiple critical stages, ranging from refining to the distribution of final products. Within this complex system, pipeline transportation serves as a fundamental pillar of modern energy logistics, providing a cost-effective and safe means of transporting large volumes of liquid fuel over long distances to downstream markets~\citep{liao2022innovations,han2024double}. In contemporary operations, different grades of fuel—such as diesel and gasoline—are pumped sequentially into the pipeline without any physical barriers between them~\citep{mostafaei2021efficient,sidki2025multiproduct}. The continuous migration of these batches introduces highly dynamic hydraulic changes throughout the system~\citep{zhou2019hybrid}. Therefore, under the strict premise of satisfying these complex hydraulic constraints, the core objective of pipeline scheduling is to safely and punctually meet downstream market demand while simultaneously minimizing operational costs~\citep{zhang2017hybrid}.

Over the past few decades, detailed scheduling optimization for multiproduct pipelines has been widely studied across different network structures and specific operating targets. The hydraulic environment is intrinsically complex, involving highly nonlinear relationships among batch interface positions, flow rates, and pressures~\citep{lin2025optimizing}. Consequently, the detailed scheduling of multiproduct pipelines is typically formulated as a mixed-integer nonlinear programming (MINLP) problem, where the primary computational challenge lies in dealing with severe model nonconvexity~\citep{cafaro2015optimization}. To alleviate this, a widely adopted practice is to linearize the non-convex constraints to form a Mixed-Integer Linear Programming (MILP) formulation. However,  While this improves tractability, it sacrifices hydraulic accuracy, and the resulting MILP models still necessitate extensive binary variables to encode spatial and temporal operational decisions. As key scale parameters increase, these variables grow exponentially, inevitably triggering a combinatorial explosion in the solution space~\citep{liao2026optimizing}. With the rapid development of data science and artificial intelligence, solving such complex scheduling optimization problems with significantly lower time costs has become possible~\citep{tu2024machine}. Therefore, this research aims to establish a rigorous mathematical modeling approach and investigate effective reinforcement learning algorithms for optimizing multiproduct pipeline scheduling.

\subsection{Related work}
To mathematically formulate this scheduling process, researchers must primarily decide how to represent time in their models. Discrete-time representations divide the entire scheduling horizon into uniform, fixed-length time intervals. \citet{rejowski2003scheduling} formulated a discrete-time MILP model as a representative approach for single pipeline scheduling problems, and \citet{meira2021solution} proposed a new discrete-time MILP model to address the long-term scheduling of pumping and delivery operations in straight multiproduct pipeline systems connecting multiple sources to multiple destinations. While predefined discrete intervals simplify basic constraint formulation, they force a severe trade-off between coarse intervals that compromise hydraulic accuracy and fine intervals that trigger the curse of dimensionality~\citep{liao2018milp}. Contrary to discrete models, continuous-time MILP representations have proven to be an efficient way to find the optimal transportation plan for single pipelines~\citep{cafaro2004optimal}. By modeling operational event timings as continuous variables rather than fixed-duration boundaries, this representation allows the model to track batch movements while generally yielding formulations with smaller dimensions~\citep{rejowski2008novel}. \citet{abdellaoui2021multi} developed a continuous-time MILP model to comprehensively optimize complex pipeline networks. \citet{mostafaei2021holistic} proposed a holistic continuous-time approach to rigorously enforce strict operational constraints, such as forbidden product sequences and dynamic inventory. Hybrid-time representations merge discrete and continuous frameworks by establishing discrete time segments, which allows the model to evaluate macroscopic states over predefined intervals while continuously tracking microscopic hydraulic events. Utilizing this logic, \citet{zhou2019hybrid} proposed a hybrid-time MILP model that accurately captures dynamic variations in pipeline frictional pressure drops during multi-batch sequential transportation.  

Even with these advanced formulations, solving the resulting models remains computationally daunting. To enhance solution efficiency, many researchers have linearized nonlinear equations associated with pressure, such as pipeline friction and pump characteristic equations~\citep{liao2018milp}, or used flow constraints instead of pressure constraints to avoid the need for nonlinear terms~\citep{liao2018milpb}. These linearized models are subsequently solved using standard exact solvers such as CPLEX or Gurobi. However, this practice faces a twofold limitation. Mathematically, imposing nonlinear physical constraints as linear bounds inherently shrinks the feasible region and degrades hydraulic accuracy, failing to guarantee a true global optimum for the actual system~\citep{lin2025optimizing}. Computationally, even linearized models face severe practical limitations. The extensive binary variables required for rigorous batch tracking impose a massive computational burden, making it highly challenging for exact solvers to tackle practical-scale scheduling problems within an acceptable timeframe~\citep{xu2021efficient,mirhassani2011scheduling}. To overcome these limitations and expedite the solution process, researchers have developed various decomposition frameworks. Two-stage decomposition algorithms have been widely used to systematically decouple the sequencing of batches and time events from the continuous calculation of detailed delivery volumes and flow rates~\citep{yan2021effective,zhang2017hybrid}. \citet{meira2021solution} and \citet{liao2026optimizing} applied rolling-horizon frameworks and multi-stage decomposition to slice long-term tasks into manageable subproblems, which not only accommodate dynamic demand updates but also drastically accelerate the solution process. Concurrently, heuristic and meta-heuristic approaches have been extensively explored. \citet{chen2017optimization} applied a simulated annealing algorithm to optimize the schedule of multiproduct pipelines, while \citet{bamoumen2023efficient} designed a greedy randomized adaptive search procedure (GRASP) to sequence batches and product volumes rapidly. Despite these improvements, heuristic rules are usually highly targeted to a specific scheduling issue, resulting in poor generality when applied to other scheduling systems with slightly different operational requirements~\citep{liao2019data}. With the development of data science and machine learning, data-driven methods have been introduced to enrich the solution approaches~\citep{liao2019data,haoran2018self}. These methods attempt to accelerate the initial search by mining historical schedules, assuming that analogous operational conditions will yield similar solutions. Consequently, they suffer from poor generalizability and fail to provide valid scheduling guidance in unprecedented scenarios lacking historical precedent.   

In recent years, reinforcement learning (RL) has been recognized as a powerful approach for optimal sequential decision-making. Benefiting from deep neural networks, deep reinforcement learning (DRL) can handle decision-making problems with large state and action spaces, achieving tremendous success in complex domains such as robotics and games~\citep{mnih2015human,kaufmann2023champion}. More notably, DRL has proven highly successful in solving analogous network operation and scheduling tasks, such as real-time pump control in water distribution networks~\citep{pei2025real}, network-constrained economic dispatch in large-scale power grids~\citep{chen2025knowledge}, continuous-action resource dispatch in isolated microgrids~\citep{dominguez2023twin}, and compressor operation as well as supply reliability optimization in natural gas pipelines~\citep{liu2026physics, fan2022systematic}. These successes across diverse networked infrastructures demonstrate the broad potential of DRL for complex operation and scheduling problems. In multiproduct pipelines specifically, DRL has so far been limited to pump scheduling~\citep{shao2025efficient}, while the detailed batch scheduling problem remains largely unexplored. However, directly applying standard DRL algorithms to this problem introduces a critical technical barrier regarding the discrete-continuous hybrid action space. Standard algorithms often struggle in complex engineering applications where an agent must simultaneously output discrete and continuous variables, a challenge recently highlighted in hybrid vehicle energy management~\citep{liu2024deep}. Multiproduct pipeline scheduling poses a comparable challenge because the agent must simultaneously make discrete operational decisions while adjusting continuous flow rates. Standard single-actor algorithms address this by treating both output types as a single vector. Inevitably, such rigid coupling neglects conditional dependencies among variables, leading the neural network to compute gradients for continuous parameters even when the corresponding discrete decisions render them operationally inactive. Consequently, forcing the agent to evaluate these invalid actions during exploration introduces severe noise into the training process, significantly hindering algorithmic convergence and reducing sample efficiency~\citep{li2025spad}. To address this structural limitation, recent studies demonstrate that spatially decoupling multidimensional actions effectively enhances policy flexibility and learning efficiency~\citep{nie2025select}.  

Inspired by these advancements, this study proposes a novel action-decoupled deep reinforcement Learning framework to optimize the detailed scheduling of multiproduct pipelines. To the best of our knowledge, this is the first work to apply deep reinforcement learning to the detailed batch scheduling of single-source, multiple-destination multiproduct pipelines. Experimental results show that the proposed framework yields high-quality detailed schedules for multiproduct pipelines with substantially reduced computation cost. The contributions of this paper are summarized as follows:
\begin{itemize}
    \item We propose a hybrid-time MINLP model for the detailed scheduling of multiproduct pipelines that preserves the full hydraulic nonlinearity rather than relying on the linearization commonly adopted in prior work. By tracking exact batch arrival times, the model enforces rigorous batch tracking and accurate hydraulic loss calculation, and formulates scheduling as a multi-objective problem that jointly minimizes delivery deviation and flow rate fluctuation, thereby providing a high-fidelity formulation without sacrificing solution accuracy.
    \item We translate the detailed scheduling process into a semi-Markov decision process (SMDP)~\citep{sutton1998reinforcement}. This framework natively accommodates flexible operational durations, enabling the agent to maintain specific scheduling schemes over variable time intervals. This SMDP reformulation thus enables a deep reinforcement learning approach to be applied directly to the detailed scheduling of multiproduct pipelines.
    \item To address the hybrid action space dilemma, we design a novel cascaded dual-actor architecture built on proximal policy optimization (PPO)~\citep{schulman2017proximal} that decouples operating duration from flow rates. This logical decoupling of discrete and continuous actions eliminates gradient interference from invalid actions, thereby enhancing sample efficiency and training stability.
\end{itemize}

The remainder of this paper is organized as follows. Section~\ref{sec2} describes the detailed scheduling problem of multiproduct pipelines and states the underlying assumptions. Section~\ref{sec3} formulates the hybrid-time MINLP model. Section~\ref{sec4} translates the scheduling process as an SMDP and presents the proposed action-decoupled SMDP-PPO framework. Section~\ref{sec5} reports the case study on a real-world pipeline, and Section~\ref{sec6} concludes the paper and discusses future work.

\section{Problem description}\label{sec2}
\begin{figure}[pos=htbp]
    \centering
    \includegraphics[width=0.9\textwidth]{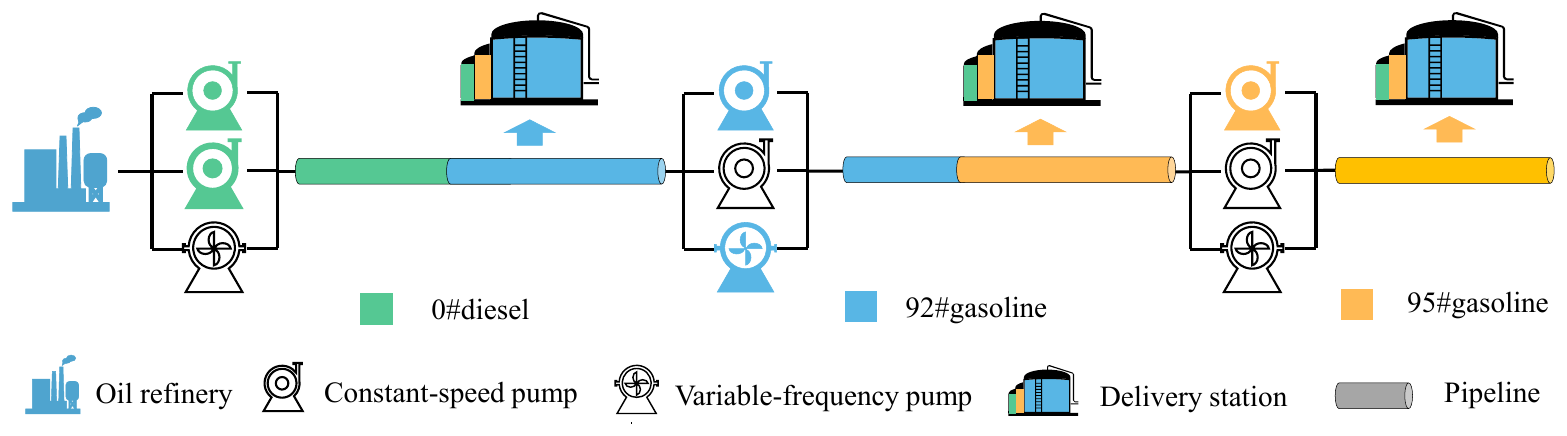}
    \caption{Schematic of sequential batch transportation in a multiproduct pipeline}
    \label{fig:trans}
\end{figure}

A multiproduct pipeline system sequentially transports several refined products from a refinery to downstream delivery station. As illustrated in Fig.~\ref{fig:trans}, the system consists of an initial injection station, a series of pipeline segments, intermediate pump stations, and delivery stations. The products are injected sequentially, without physical separators, and travel forward as distinct batches. Pump stations are strategically placed along the route to provide the necessary pressure to overcome frictional losses and elevation differences, thereby maintaining continuous flow. As batches advance through the pipeline, the intermediate and terminal delivery stations withdraw the required volume from the corresponding batches to meet local market demand.

Based on downstream market demands and injection plans, detailed pipeline scheduling determines the specific execution scheme for each operational period, determining the exact delivery volumes and delivery flow rates. In this study, the optimization model comprehensively addresses three key aspects: minimizing supply-demand deviations, maintaining flow rate stability, and accurately calculating dynamic hydraulic losses. The objective is to minimize delivery deviation while also pursuing flow rate stability. While precise delivery ensures supply reliability, maintaining stable flow rates effectively minimizes frictional pressure drop and thereby helps curb the associated pumping energy cost. It also improves scheduling practicability by reducing the operator workload from frequent valve adjustments and pump switching~\citep{chen2017optimizing}. To ensure the feasibility of the schedule, the proposed model enforces exact batch tracking by incorporating the exact arrival times of batches at each station. This precise tracking enables accurate calculation of hydraulic loss, ensuring that all operational decisions comply with stringent hydraulic safety limits during the batch sequential transportation process~\citep{zhou2019hybrid}.

To explicitly formulate the pipeline detailed scheduling problem, the given parameters and the decision variables to be determined are formally defined as follows.

Given:
\begin{itemize}
    \item Pipeline data: Length, inner diameter, and flow rate limits of each pipeline segment.
    \item Station data: Elevation of each station, injection and delivery flow rate limits, max head of pump station.
    \item Production data: Physical properties of products (density, viscosity), initial batch states, predetermined batch injection volumes and sequence.
    \item Demand plan: Target delivery volumes for each product at respective delivery stations.
\end{itemize}

Determine:
\begin{itemize}
    \item Injection schedule: Injection flow rate and execution durations.
    \item Delivery schedule: Delivery flow rate and execution durations.
\end{itemize}

To effectively solve this model, the following reasonable simplifications and assumptions are made:
\begin{itemize}
    \item Refined products are treated as incompressible fluids with constant physical properties, completely disregarding any temperature- or pressure-induced variations.
    \item Within any time interval, the continuous variables, including station injection/delivery flow rate, remain strictly steady.
    \item  Fluid flow in all segments is strictly assumed to be hydraulically smooth.
    \item Contamination between adjacent batches is modeled as a sharp zero-volume boundary, thereby neglecting the complex tracking of mixed-oil sections.
    \item Tank inventory constraints are excluded, assuming all stations have sufficient capacity to store the scheduled products.
\end{itemize}

\section{Mathematical model}\label{sec3}
In this section, we formulate the detailed scheduling problem for a multiproduct pipeline with a single injection source and several delivery stations as a hybrid-time MINLP model. The objective minimizes delivery deviations and flow rate fluctuations, subject to time node, batch tracking, delivery, flow rate, and pressure constraints. The Appendix~\ref{app:nomenclature} contains a complete nomenclature.

\subsection{Objective function}
The multiproduct pipeline detailed scheduling is developed as a multi-objective optimization problem that minimizes the delivery deviation ($f_1$) and the flow rate fluctuation ($f_2$), as shown in Eq.~\eqref{eq:objective}.
\begin{equation}
\min f=f_1+f_2\label{eq:objective}
\end{equation}
The delivery deviation $f_1$ is defined as the absolute difference between the planned demand and the actual volumes delivered at each delivery station.
\begin{equation}
f_1=\sum_i \sum_o \left|V^x_{i, o}-V^d_{i, o}\right| \label{eq:objective1}
\end{equation}

The stability objective $f_2$ suppresses operational fluctuations by penalizing the variation in flow rates between consecutive time intervals for each pipeline segment $i$.
\begin{equation}
f_2=\sum_{k=2}^{k_{\max} -1} \sum_{i=1}^{i_{\max} -1} \lambda_{i} \cdot\left|Q^p_{k, i}-Q^p_{k-1, i}\right|\label{eq:fluctuations}
\end{equation}
A weight factor $\lambda_{i}$ is applied to discourage fluctuations in flow rate in pipeline segments with larger volume capacities, as flow variations in such segments lead to higher frictional pressure, as expressed in Eq.~\eqref{eq:lambda}.
\begin{equation}
\lambda_{i}=\frac{V^s_{i+1}-V^s_i}{V^s_{i_{\max} }} \label{eq:lambda}
\end{equation}

\subsection{Time node constraints}
The scheduling horizon is represented by a sequence of time nodes $t_k$ accumulated through Eq. \eqref{eq:step_acc}, starting from $t_1 = 0$.
\begin{equation}
t_{k+1} = t_k + \Delta t_k \quad \forall k \in K \setminus \{k_{\max}\} \label{eq:step_acc}
\end{equation}

A binary variable $B_{k+1}$ equals 1 if a batch head interface reaches a station at node $k+1$ and 0 otherwise, as determined by Eqs. \eqref{eq:event1} and \eqref{eq:event2}.
\begin{equation}
B_{k+1} \ge S^{head}_{k+1,i,j} - S^{head}_{k,i,j} \quad \forall k \in K \setminus \{k_{\max}\}, i \in I, j \in J \label{eq:event1}
\end{equation}
\begin{equation}
B_{k+1} \le \sum_{i \in I} \sum_{j \in J} \left( S^{head}_{k+1,i,j} - S^{head}_{k,i,j} \right) \quad \forall k \in K \setminus \{k_{\max}\} \label{eq:event2}
\end{equation}

The step $\Delta t_k$ is fixed to the base interval when no batch head interface reaches a station, and is released to a variable value within $(0,1]$ when a batch head arrives, as enforced by Eqs. \eqref{eq:step_ub} and \eqref{eq:step_lb}, where $M$ is a big-M constant and $\varepsilon$ is a small positive constant that prevents a zero-length step.
\begin{equation}
1 - B_{k+1} M \le \Delta t_k \le 1 \quad \forall k \in K \setminus \{k_{\max}\} \label{eq:step_ub}
\end{equation}
\begin{equation}
\Delta t_k \ge \varepsilon B_{k+1} + (1 - B_{k+1}) \quad \forall k \in K \setminus \{k_{\max}\} \label{eq:step_lb}
\end{equation}

\subsection{Batch tracking constraints}
The initial positions of all batches must be defined. For batches already inside the pipeline, the initial volumetric coordinate is determined by the known initial position $V_j^{init}$, as expressed in Eq.~\eqref{eq:oldbatch}. For new batches to be injected, the starting position is assigned a negative value calculated by Eq.~\eqref{eq:newbatch}, which represents the cumulative planned injection volumes $V_{j'}^{inj}$ of all preceding new batches.
\begin{equation}
V_{1, j}^b = V_j^{init} \quad \forall j \in J^{old}\label{eq:oldbatch}
\end{equation}
\begin{equation}
V_{1, j}^b = - \sum_{j'=1}^{j-1} V_{j'}^{inj} \quad \forall j \in J^{new}\label{eq:newbatch}
\end{equation}

The binary indicator $S^{head}_{k,i,j}$ is introduced as a fundamental tracking variable. As defined by the big-M relaxation in Eqs.~\eqref{eq:Vbig} and~\eqref{eq:Vsmall}, $S^{head}_{k,i,j}$ equals 1 if the head interface of batch $j$ has passed station $i$ at time node $k$, and 0 otherwise. Eqs.~\eqref{eq:SI} and~\eqref{eq:SJ} preserve the physical order by ensuring that each batch passes stations sequentially and that all batches maintain their relative positions.
\begin{equation}
V^b_{k,j} - V^s_i \ge (S^{head}_{k,i,j} - 1)M \quad \forall k \in K, i \in I, j \in J\label{eq:Vbig}
\end{equation}
\begin{equation}
V^b_{k,j} - V^s_i \le S^{head}_{k,i,j} M \quad \forall k \in K, i \in I, j \in J\label{eq:Vsmall}
\end{equation}
\begin{equation}
S^{head}_{k,i,j} \ge S^{head}_{k,i+1,j} \quad \forall k \in K, i \in I \setminus \{i_{\max}\}, j \in J\label{eq:SI}
\end{equation}
\begin{equation}
S^{head}_{k,i,j} \ge S^{head}_{k,i,j+1} \quad \forall k \in K, i \in I, j \in J \setminus \{j_{\max}\}\label{eq:SJ}
\end{equation}

When a batch head interface reaches a station, i.e., $\left(S_{k+1, i, j}^{\text {head }}-S_{k, i, j}^{\text {head }}\right)=1$, the time node constraints in Eqs.~\eqref{eq:event1}-\eqref{eq:step_lb} release the step length $\Delta t_k$ to a variable value, while Eq.~\eqref{eq:Vequal} coupled with Eq.~\eqref{eq:Vbig} forces the coordinate $V_{k+1, j}^b$ to coincide with the station coordinate $V_i^s$. Through the volume update in Eq.~\eqref{eq:vupdate}, this alignment determines the exact $\Delta t_k$ and hence the precise arrival time $t_{k+1}$ via Eq.~\eqref{eq:step_acc}.
\begin{equation}
V^b_{k+1,j} - V^s_i \le \left( 1 - S^{head}_{k+1,i,j} + S^{head}_{k,i,j} \right) M \quad \forall k \in K \setminus \{k_{\max}\}, i \in I, j \in J\label{eq:Vequal}
\end{equation}

Product delivery from batch $j$ at station $i$ is permitted only when the batch covers the station, a condition captured by the binary variable $S^{cover}_{k,i,j}$ in Eqs.~\eqref{eq:scover} and~\eqref{eq:scovermax}, which indicates that the head interface has passed the station while the tail interface remains upstream.
\begin{equation}
S^{cover}_{k,i,j} = S^{head}_{k,i,j} - S^{head}_{k,i,j+1} \quad \forall k \in K, i \in I, j \in J \setminus \{j_{\max}\}\label{eq:scover}
\end{equation}
\begin{equation}
S^{cover}_{k,i,j_{\max}} = S^{head}_{k,i,j_{\max}} \quad \forall k \in K, i \in I\label{eq:scovermax}
\end{equation}

The volumetric coordinate $V^b_{k,j}$ is updated in Eq. \eqref{eq:vupdate} by adding the volumetric displacement during the time interval, which is calculated as the difference between the source injection and all upstream deliveries. 
\begin{equation}
V^b_{k+1,j} = V^b_{k,j} + \left( Q^{in}_k - \sum_{i \in I} S^{head}_{k,i,j} \cdot Q^d_{k,i} \right) \Delta t_k\ \quad \forall k \in K \setminus \{k_{\max}\}, j \in J\label{eq:vupdate}
\end{equation}

As hydraulic loss varies with oil properties, the spatial relationship between batches and pipeline segments must be precisely determined. To facilitate this calculation, the occupancy of a batch within a segment is described by four mutually exclusive binary states. Since the last batch $j_{max}$ has no succeeding batch and thus no tail interface within the horizon, the head indicator of the nonexistent succeeding batch is set to zero, i.e., $S_{k, i, j_{\max }+1}^{head}=0$ for all $k \in K, i \in I$, so that the following definitions apply uniformly to all batches.

As defined in Eqs. \eqref{eq:thru1}–\eqref{eq:thru3}, the binary variable $Z^{thru}_{k,i,j}$ equals 1 if batch $j$ completely passes through segment $i$, which occurs when its head interface has passed station $i+1$ while its tail interface has not yet reached station $i$.
\begin{equation}
Z^{thru}_{k,i,j} \le S^{head}_{k,i+1,j} \quad \forall k \in K, i \in I \setminus \{i_{\max}\}, j \in J \label{eq:thru1}
\end{equation}
\begin{equation}
Z^{thru}_{k,i,j} \le 1 - S^{head}_{k,i,j+1} \quad \forall k \in K, i \in I \setminus \{i_{\max}\}, j \in J \label{eq:thru2}
\end{equation}
\begin{equation}
Z^{thru}_{k,i,j} \ge S^{head}_{k,i+1,j} - S^{head}_{k,i,j+1} \quad \forall k \in K, i \in I \setminus \{i_{\max}\}, j \in J \label{eq:thru3}
\end{equation}

As defined in Eqs. \eqref{eq:head1}–\eqref{eq:head3}, the binary variable $Z^{H\_in}_{k,i,j}$ equals 1 if the head interface of batch $j$ is inside segment $i$ and its tail interface has not yet reached station $i$.
\begin{equation}
Z^{H\_in}_{k,i,j} \le S^{head}_{k,i,j} - S^{head}_{k,i+1,j} \quad \forall k \in K, i \in I \setminus \{i_{\max}\}, j \in J \label{eq:head1}
\end{equation}
\begin{equation}
Z^{H\_in}_{k,i,j} \le 1 - S^{head}_{k,i,j+1} \quad \forall k \in K, i \in I \setminus \{i_{\max}\}, j \in J  \label{eq:head2}
\end{equation}

\begin{equation}
Z^{H\_in}_{k,i,j} \ge (S^{head}_{k,i,j} - S^{head}_{k,i+1,j}) - S^{head}_{k,i,j+1} \quad \forall k \in K, i \in I \setminus \{i_{\max}\}, j \in J \label{eq:head3}
\end{equation}

As defined in Eqs. \eqref{eq:tail1}–\eqref{eq:tail3}, the binary variable $Z^{T\_in}_{k,i,j}$ equals 1 if the tail interface of batch $j$ is inside segment $i$ and its head interface has passed station $i+1$.
\begin{equation}
Z^{T\_in}_{k,i,j} \le S^{head}_{k,i+1,j} \quad \forall k \in K, i \in I \setminus \{i_{\max}\}, j \in J \label{eq:tail1}
\end{equation}
\begin{equation}
Z^{T\_in}_{k,i,j} \le S^{head}_{k,i,j+1} - S^{head}_{k,i+1,j+1} \quad \forall k \in K, i \in I \setminus \{i_{\max}\}, j \in J \label{eq:tail2}
\end{equation}
\begin{equation}
Z^{T\_in}_{k,i,j} \ge S^{head}_{k,i+1,j} + (S^{head}_{k,i,j+1} - S^{head}_{k,i+1,j+1}) - 1 \quad \forall k \in K, i \in I \setminus \{i_{\max}\}, j \in J \label{eq:tail3}
\end{equation}

As defined in Eqs. \eqref{eq:all1}–\eqref{eq:all3}, the binary variable $Z^{all\_in}_{k,i,j}$ equals 1 if both the head and tail interfaces of batch $j$ are inside segment $i$.
\begin{equation}
 Z^{all\_in}_{k,i,j} \le S^{head}_{k,i,j} - S^{head}_{k,i+1,j} \quad \forall k \in K, i \in I \setminus \{i_{\max}\}, j \in J \label{eq:all1}
\end{equation}
\begin{equation}
 Z^{all\_in}_{k,i,j} \le S^{head}_{k,i,j+1} - S^{head}_{k,i+1,j+1} \quad \forall k \in K, i \in I \setminus \{i_{\max}\}, j \in J  \label{eq:all2}
\end{equation}
\begin{equation}
Z^{all\_in}_{k,i,j} \ge (S^{head}_{k,i,j} - S^{head}_{k,i+1,j}) + (S^{head}_{k,i,j+1} - S^{head}_{k,i+1,j+1}) - 1 \quad \forall k \in K, i \in I \setminus \{i_{\max}\}, j \in J \label{eq:all3}
\end{equation}

\subsection{Delivery constraints}
Due to the technical limits of the measuring equipment and the limited receiving capacity of the station, the delivery flow rate $Q^d_{k,i}$ is constrained by Eq. \eqref{eq:flow_limit} to be zero or operate within its minimum threshold and maximum capacity.
\begin{equation}
S^{d}_{k,i} \cdot Q^{d,\min}_i \le Q^d_{k,i} \le S^{d}_{k,i} \cdot Q^{d,\max}_i \quad \forall k \in K, i \in I\label{eq:flow_limit}
\end{equation}

Summing the delivery volumes across all intervals where product $o$ covers station $i$ yields the total delivered volume, as expressed in Eq. \eqref{eq:total_vol}.
\begin{equation}
V^d_{i, o} = \sum_{k=1}^{k_{\max}-1} \sum_{j \in J_o} \left( S^{cover}_{k,i,j} \cdot Q^d_{k,i} \right)  \Delta t_k \quad \forall i \in I, o \in O\label{eq:total_vol}
\end{equation}

\subsection{Flow rate constraints}
The segment flow rate $Q^p_{k, i}$ equals the adjacent upstream segment’s flow rate minus the delivery at the connecting station, as defined in Eq. \eqref{eq:flow2}. For the first segment, it equals the injection at the initial station in Eq. \eqref{eq:flow1}.
\begin{equation}
Q^p_{k, 1} = Q^{in}_k \quad \forall k \in K\label{eq:flow1}
\end{equation}
\begin{equation}
Q^p_{k, i} = Q^p_{k, i-1} - Q^d_{k, i} \quad \forall k \in K, i \in I \setminus \{1,i_{\max }\}\label{eq:flow2}
\end{equation}

Furthermore, to ensure transportation efficiency and equipment safety, the flow rate in each pipeline segment must be maintained within a specific allowable range, as specified by Eq. \eqref{eq:segment_flow}.
\begin{equation}
Q^{p,\min}_i \le Q^p_{k, i} \le Q^{p,\max}_i \quad \forall k \in K, i \in I \setminus\{i_{\max }\}\label{eq:segment_flow}
\end{equation}

\subsection{Pressure constraints}
The nonlinear relationship between the flow rate and the pipeline frictional pressure is typically calculated using the empirical Leibenzon formula, which is widely used for friction loss calculations in multiproduct pipelines in China~\citep{zhang2017hybrid,lin2025optimizing}. Assuming the oil flow regime stays within the hydrodynamic smooth zone ($\beta=0.0246, m=0.25$), the basic friction head loss $h_l$ is defined by:
\begin{equation}
h_l=\beta \frac{Q^{2-m} \nu^m}{d^{5-m}} L.
\end{equation}

\begin{figure}[pos=htbp]
    \centering
    \includegraphics[width=0.9\textwidth]{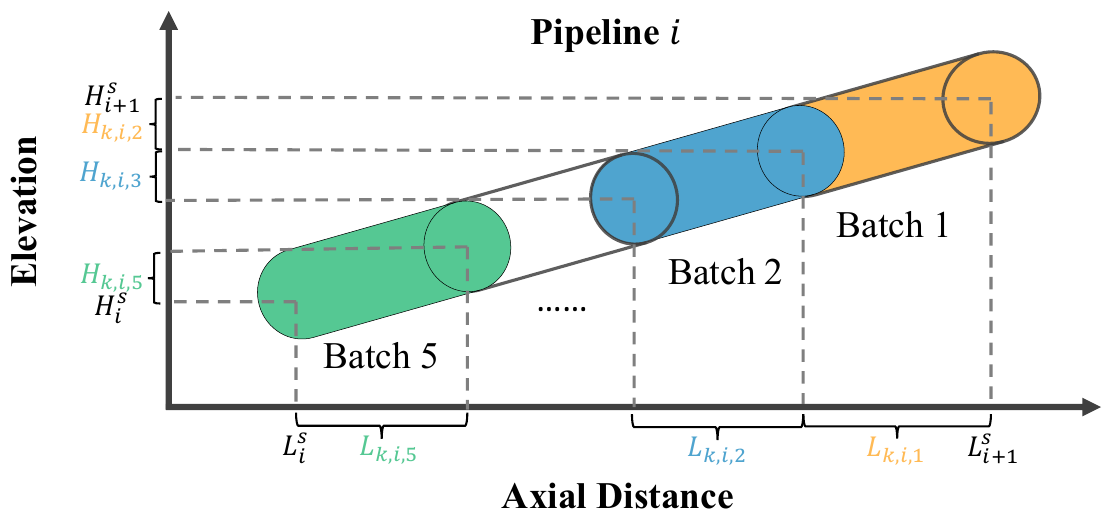}
    \caption{Schematic diagram of batch lengths and elevation differences within pipeline segment $i$.}
    \label{fig:pipeline_profile}
\end{figure}

Fig. \ref{fig:pipeline_profile}. illustrates the schematic for pressure drop calculation in a multiproduct pipeline. As oil products with different physical properties move sequentially, the frictional and elevation-related pressure drops change dynamically. At each time interval $k$, the total pressure required for segment $i$ is determined by summing the effects of all batches within that segment; specifically, the frictional pressure drop $P^F_{k,i}$ and elevation-related pressure drop $P^H_{k,i}$ are defined by Eq. \eqref{eq:PF_gen} and Eq. \eqref{eq:PH_gen}, respectively.
\begin{equation}
P^F_{k,i} = \frac{g \beta (Q^p_{k,i})^{2-m}}{d_i^{5-m}} \left[ \rho_1 \nu_1^m L_{k,i,1} + \rho_2 \nu_2^m L_{k,i,2} + \dots + \rho_{j_{\max}} \nu_{j_{\max}}^m L_{k,i,j_{\max}} \right] \quad \forall k \in K, i \in I \setminus \{i_{\max }\}\label{eq:PF_gen}
\end{equation}
\begin{equation}
P^H_{k,i} = g \left[ \rho_1 H_{k,i,1} + \rho_2 H_{k,i,2} + \dots + \rho_{j_{\max}} H_{k,i,j_{\max}} \right] \quad \forall k \in K, i \in I \setminus \{i_{\max }\}\label{eq:PH_gen}
\end{equation}

Here, the exact physical length $L_{k,i,j}$ and the elevation difference $H_{k,i,j}$ spanned by batch $j$ inside segment $i$ are precisely determined by the four mutually exclusive binary topological variables ($Z^{thru}, Z^{H\_in}, Z^{T\_in}, Z^{all\_in}$) formulated in the batch tracking section. By combining these with the volumetric coordinates, the detailed pressure drop formulas are derived:
\begin{equation}
P^F_{k,i} = \frac{4 g \beta (Q^p_{k,i})^{2-m}}{\pi d_i^{7-m}} \sum_{j=1}^{j_{\max}} \left\{ \rho_j \nu_j^m \left[ Z^{thru}_{k,i,j}(V^s_{i+1} - V^s_i) + Z^{H\_in}_{k,i,j}(V^b_{k,j} - V^s_i) + Z^{T\_in}_{k,i,j}(V^s_{i+1} - V^b_{k,j+1}) + Z^{all\_in}_{k,i,j}(V^b_{k,j} - V^b_{k,j+1}) \right] \right\},
\end{equation}
\begin{equation}
P^H_{k,i} = \frac{g (E_{i+1} - E_i)}{V^s_{i+1} - V^s_i} \sum_{j=1}^{j_{\max}} \left\{ \rho_j \left[ Z^{thru}_{k,i,j}(V^s_{i+1} - V^s_i) + Z^{H\_in}_{k,i,j}(V^b_{k,j} - V^s_i) + Z^{T\_in}_{k,i,j}(V^s_{i+1} - V^b_{k,j+1}) + Z^{all\_in}_{k,i,j}(V^b_{k,j} - V^b_{k,j+1}) \right] \right\}.
\end{equation}
The sum $P_{k, i}^F+P_{k, i}^H$ gives the total pressure drop, i.e., the hydraulic loss, of segment $i$ at node $k$. Under the hybrid time framework, the transportation flow rate remains constant throughout any interval $(t_k, t_{k+1})$, during which no batch interfaces are permitted to enter or leave the pipeline segments. Consequently, the total pipeline pressure drop exhibits a strictly linear monotonic change over time during this interval~\citep{zhou2019hybrid}. To ensure safety, the cumulative pressure demand up to node $i$ must stay within the total upstream supply capacity. Due to pressure linearity, enforcing this constraint at both interval endpoints ($t_k$ and $t_{k+1}$) guarantees compliance at any intermediate moment, as formulated in Eqs. \eqref{eq:p_limit1} and \eqref{eq:p_limit2}:
\begin{equation}
\sum_{w=1}^i \left( P^F_{k,w} + P^H_{k,w} \right) \le \sum_{w=1}^i P^{\max}_w \quad \forall k \in K \setminus \{k_{\max}\}, i \in I \setminus \{i_{\max }\},\label{eq:p_limit1}
\end{equation}
\begin{equation}
\sum_{w=1}^i \left( P^F_{k+1,w} + P^H_{k+1,w} \right) \le \sum_{w=1}^i P^{\max}_w \quad \forall k \in K \setminus \{k_{\max}\}, i \in I \setminus \{i_{\max }\}.\label{eq:p_limit2}
\end{equation}

\section{Action-decoupled deep reinforcement learning}\label{sec4}
In this section, we propose an action-decoupled deep reinforcement learning framework to optimize the detailed scheduling of a multiproduct pipeline. This framework is based on the semi-Markov decision process (SMDP) paradigm to enhance flow rate stability ($f_2$) through a sequence of "macro-actions". Each macro-action represents a stable operating state maintained until an asynchronous decision node is triggered, either by the expiration of a planned duration or by critical events. The following subsections present the formal SMDP formulation, the proposed action-decoupled architecture, and the PPO-based optimization procedure.

\subsection{SMDP formulations for multiproduct pipeline scheduling}
The scheduling problem is first mathematically structured as a semi-Markov decision process (SMDP) to recast the scheduling problem in a form amenable to reinforcement learning. The SMDP is formally defined by a tuple $\langle \mathcal{S}, \mathcal{A}, \mathcal{R}, \mathcal{P}, \mathcal{T}, \gamma \rangle$, where $\mathcal{S}$ is the state space, $\mathcal{A}$ is the action space, $\mathcal{R}$ is the reward function, $\mathcal{P}(s', \tau \mid s, a)$ denotes the transition dynamics, and $\gamma \in (0, 1]$ is the discount factor. The transition dynamics $\mathcal{P}(s', \tau \mid s, a)$ represent the probability of transitioning to state $s'$ after an actual execution duration $\tau \in \mathcal{T}$. The objective is to learn an optimal policy $\pi^*$ that maximizes the expected discounted rewards
$G = \mathbb{E} \left[ \sum_{n=0}^{L-1} \gamma^{t_n} R_{t_n} \right]$where $L$ is the number of decision stages and $t_n = \sum_{l=0}^{n-1} \tau_l$ is the cumulative number of steps at stage $n$ (with $t_0 = 0$). Here, $\tau_n \in \mathcal{T}$ represents the actual execution duration of the $n$-th macro-action. To maintain temporal consistency, the stage reward $R_{t_n}$ is defined as the discounted accumulation of instantaneous rewards $R(\cdot)$ over the actual duration $\tau_n$:
$R_{t_n} = \sum_{m=0}^{\tau_n-1} \gamma^{m} R(s_{t_n+m}, a_{t_n})$. Here, $m$ denotes the discrete step index within the $n$-th macro-action, and $s_{t_n+m}$ represents the environmental state at step $t_n+m$.

The components of the SMDP are specified as follows:

\noindent \textbf{1) State space ($\mathcal{S}$):}
The state at decision stage $n$ is represented by a feature vector $s_{t_n} = \{ V^b_{t_n}, \Delta V_{t_n}, \rho_{t_n}, v_{t_n}, a_{t_{n-1}} \}$. This vector encompasses the volumetric coordinates of all batch head interfaces $V^b_{t_n}$, the densities $\rho_{t_n}$ and kinematic viscosities $v_{t_n}$ of the products currently passing through each station, the remaining delivery demand $\Delta V_{t_n}$ for each station and product type, and the macro-action $a_{t_{n-1}}$ executed in the previous stage.
    
\noindent \textbf{2) Action space ($\mathcal{A}$):}
Each macro-action $a_{t_n} = \{ d_n, q_{t_n} \}$ in the hybrid action space $\mathcal{A}$ represents a stable operating state. Here, $d_n \in \mathcal{T}$ is the discrete planned duration, while the continuous vector $q_{t_n}$ specifies both the injection flow rate $Q^{in}_{t_n}$ at the initial station and the delivery flow rates $ Q^ {d}_{{t_n},i}$ at all downstream delivery stations.

\noindent \textbf{3) Reward function ($\mathcal{R}$):}
The reward function operationalizes the multi-objective optimization as RL feedback, providing the agent with progressive guidance by penalizing delivery deviations, operational instability, and safety violations. The total stage reward $R_{t_n}$ consists of an accumulated deviation component and several macro-level penalties:
\begin{equation}
R_{t_n} = \alpha R_{de} + \beta \cdot R_{sta} + \eta \cdot R_{p},
\end{equation}
where $\alpha, \beta$ and $\eta$ are weight factors used to balance the operational objectives. Each term is defined as follows:
\begin{itemize}
    \item \textbf{Delivery deviation reward ($R_{de}$):} This component reflects the delivery deviation objective $f_1$ defined in Eq. \eqref{eq:objective1}. It consists of an interval-based penalty $R_{int}$ and a terminal penalty $R_{term}$ triggered at the end of the horizon:
    \begin{equation}
     R_{de} = \sum_{m=0}^{\tau_n-1} \gamma^{m} R_{int}(s_{t_n+m}) + \mathbb{I}(done) \cdot R_{term}.
    \end{equation}
    The interval-based penalty $R_{int}$ penalizes delivery errors at each discrete step $t_n+m$ within the macro-action through a piecewise function specifically designed to enhance gradient sensitivity to minor errors:
    \begin{equation}
    R_{int}(s_{t_n+m}) = \begin{cases} -\mathcal{E}_{t_n+m}, & \text{if } \mathcal{E}_{t_n+m} > \delta \\ -\ln(\mathcal{E}_{t_n+m} + \varepsilon), & \text{if } \mathcal{E}_{t_n+m} \le \delta \end{cases},
    \end{equation}
    where $\mathcal{E}_{t_n+m}$ represents the normalized root mean square error (NRMSE) of the deviations between actual and planned delivery volumes in all stations and products at step $t_n+m$, and $\delta$ denotes an empirical threshold that balances coarse-scale and fine-scale optimization.
    The terminal penalty $R_{term}$ imposes a strict global constraint at the final step:
    \begin{equation}
    R_{term} = - \Gamma \cdot \frac{\sum_i \sum_o \left|V^x_{i,o} - V^d_{i,o}\right|}{\sum_i \sum_o V^x_{i,o}},
    \end{equation}
    where $\Gamma$ is a penalty coefficient and $\mathbb{I}(done)$ is an indicator function that equals 1 only upon the end of the scheduling horizon.
    \item \textbf{Flow rate stability reward ($R_{sta}$ ):} This component encourages steady operations, corresponding to the stability objective $f_2$ in Eq. \eqref{eq:fluctuations}. It is composed of an operational smoothness reward ($R_{sw}$) and a switching penalty ($R_{dur}$), i.e., $R_{sta}=R_{sw}+R_{dur}$, where:
    \begin{equation}
    R_{sw} = - \sum_{i=1}^{i_{\max}-1} \lambda_{i} \left| \frac{Q^p_{t_n, i} - Q^p_{t_{n-1}, i}}{Q_{\max}} \right|,
    \end{equation}
    \begin{equation}
    R_{dur} = -C_{switch}.
    \end{equation}
    
    Here $R_{sw}$ penalizes flow variations across pipeline segments between decision stages using volumetric weights $\lambda_i$, with $Q_{\max}$ serving as a normalization constant representing the maximum allowable flow rate of the pipeline. $R_{dur}$ reduces intervention frequency by imposing a switching cost to encourage longer, more stable macro-actions.
    \item \textbf{Pressure penalty ($R_{p}$):} To ensure pressure safety according to Eqs. \eqref{eq:p_limit1} and \eqref{eq:p_limit2}, a penalty $R_{p} = -\Delta P_{violation}$ is applied. Here, $\Delta P_{violation}$ represents the magnitude of the pressure violation when the hydraulic constraints are breached. Notably, such a violation is treated as a critical event, which immediately truncates the current macro-action and forces a transition to a new decision node.
\end{itemize}

\noindent \textbf{4)  Transition dynamics ($\mathcal{P}, \mathcal{T}$):}
 The transition probability $\mathcal{P}(s_{t_{n+1}}, \tau_n \mid s_{t_n}, a_{t_n})$ defines the environment's evolution over an actual duration $\tau_n$. Although the agent proposes a planned duration $d_n$, the actual duration $\tau_n$ is determined by the simulator and may be truncated by critical events, such as a batch interface arriving at a delivery station or a pressure violation.

\subsection{Action decoupling via cascaded dual-actor architecture}
Traditional policy-based DRL methods typically use a single-actor to learn a joint policy $\pi_\theta(d, q \mid s)$, where the planned scheduling duration $d$ and flow rates $q$ are generated simultaneously. In pipeline scheduling, however, these two sub-actions are strongly coupled: the appropriate flow rate $q$ depends on the duration $d$ over which it is held, so the same $q$ may be either suitable or harmful depending on $d$. Following the policy gradient theorem ~\citep{sutton1998reinforcement}, the joint objective gradient in a single-actor framework is
\begin{equation}
\nabla_\theta J(\theta) = \mathbb{E}_{\pi_\theta} [\nabla_\theta \log \pi_\theta (d, q \mid s) \cdot A(s, d, q)],
\end{equation}
where $A(s, d, q)$ is the advantage function evaluating the joint action. This leads to gradient interference: when a reasonable duration is paired with an incompatible flow rate during exploration, the low advantage of this combination drives the shared gradient to suppress the probability of the duration as well, even though the duration itself is sound. Such interference increases training instability and hampers the agent's ability to identify compatible macro-actions efficiently.

To address this, we propose an action-decoupled architecture that employs a conditional factorization of the joint policy into two serial actors:
\begin{equation}
\pi_\theta(a \mid s) = \pi_{\theta_1}(d \mid s) \cdot \pi_{\theta_2}(q \mid s, d).\label{eq:actor1}
\end{equation}

In this serial structure, Actor 1 first determines the planned duration $d$, and Actor 2 then selects the corresponding flow rates $q$ based on this duration. The decoupled optimization logic for these two actors is derived as follows:
\begin{itemize}
    \item \textbf{Duration network (Actor 1):} To derive the update rule for Actor 1, we begin with the gradient of the joint objective with respect to the duration parameters $\theta_1$. Based on the policy factorization of Eq. \eqref{eq:actor1}, the gradient is initially expressed as:
    \begin{equation}
    \begin{aligned}
    \nabla_{\theta_1} J &= \mathbb{E}_{s \sim \rho, a \sim \pi_\theta} \left[ \nabla_{\theta_1} \log \pi_\theta(a \mid s) \cdot A(s, d, q) \right] \ \\
    &= \mathbb{E}_{s \sim \rho, (d, q) \sim \pi_\theta} \left[ \nabla_{\theta_1} \log \pi_{\theta_1}(d \mid s) \cdot A(s, d, q) \right].
    \end{aligned}
    \end{equation}
    Here, the second factor $\log \pi_{\theta_2}(q \mid s, d)$ vanishes under $\nabla_{\theta_1}$ since it does not depend on $\theta_1$, leaving only the duration term. The joint expectation is decomposed into a nested form using the law of total expectation:
    \begin{equation}
    \begin{aligned}
    \nabla_{\theta_1} J &= \mathbb{E}_{s \sim \rho, d \sim \pi_{\theta_1}} \left[ \mathbb{E}_{q \sim \pi_{\theta_2}} \left[ \nabla_{\theta_1} \log \pi_{\theta_1}(d \mid s) \cdot A(s, d, q) \mid s, d \right] \right] \ \\
    &= \mathbb{E}_{s \sim \rho, d \sim \pi_{\theta_1}} \left[ \nabla_{\theta_1} \log \pi_{\theta_1} (d \mid s) \cdot \mathbb{E}_{q \sim \pi_{\theta_2}} [A(s, d, q) \mid s, d] \right].\label{eq:expectation}
    \end{aligned}
    \end{equation}
    
    Eq. \eqref{eq:expectation} shows that the duration policy $\pi_{\theta_1}$ is updated based on the marginal advantage evaluated over the flow rate distribution. By evaluating the duration $d$ based on the expected performance of subsequent flow rates, Actor 1 can determine whether a specific duration is suitable for the current multiproduct pipeline state. This approach prevents the duration policy from being misled by a single mismatched flow rate sample during exploration, ensuring more robust and stable credit assignment during the early training phase.
    \item \textbf{Flow rate network (Actor 2):} Conditioned on the duration $d$ from the first stage, the flow rate policy $\pi_{\theta_2}$ is optimized to find the most compatible flow rate control for that specific time window:
    \begin{equation}
    \nabla_{\theta_2} J = \mathbb{E}_{s, d \sim \pi_{\theta_1}, q \sim \pi_{\theta_2}} [\nabla_{\theta_2} \log \pi_{\theta_2} (q \mid s, d) \cdot A(s, d, q)].
    \end{equation}
    For Actor 2, receiving the duration $d$ as a prior condition significantly simplifies the exploration complexity, as the agent only needs to find the optimal flow rates within a given duration. Furthermore, the generated flow rates must also strictly adhere to safety boundaries. Therefore,
    to satisfy the delivery and pipeline segment flow rate constraints defined in Eqs. \eqref{eq:flow_limit} and \eqref{eq:segment_flow}, a sequential action mapping mechanism is embedded in the output layer of Actor 2. Rather than directly outputting the delivery flow rate, the network generates a raw unconstrained vector $u_{t_n} = [u_{t_n} ^{in}, u_{{t_n} ,1}^d, \dots, u_{{t_n} ,I}^d]$, which is transformed into physical actions in two steps. First, the injection flow rate $Q_{t_n}^{in}$ is mapped to strictly satisfy the capacity of the initial pipeline segment:
    \begin{equation}
    Q_{t_n}^{in} = Q^{p,\min}_1 + \sigma(u_{t_n}^{in}) \cdot (Q^{p,\max}_1 - Q^{p,\min}_1),
    \end{equation}
    where $\sigma(\cdot)$ is the sigmoid activation function. Then, the delivery flow rate $Q_{t_n,i}^d$ for each delivery station $i$ is mapped sequentially to ensure that neither the delivery capacity nor the minimum flow rate limits are violated:
    \begin{equation}
    \tilde{Q}_{t_n,i}^d = \min \left( Q^{d,\max}_i, \; Q^p_{t_n,i-1} - Q^{p,\min}_{i} \right),
    \end{equation}
    \begin{equation}
    Q_{t_n,i}^d =
    \begin{cases}
    0, & \text{if } \tilde{Q}_{t_n,i}^d < Q^{d,\min}_i \\
    Q^{d,\min}_i + \sigma(u_{t_n,i}^d) \cdot \left(\tilde{Q}_{t_n,i}^d - Q^{d,\min}_i\right), & \text{otherwise}
    \end{cases}.
    \end{equation}

    This two-step mapping restricts the action space to a strictly feasible region, ensuring that all generated flow rates are legally feasible.

\end{itemize}

This hierarchical update mechanism ensures that the environment's feedback is correctly attributed to duration selection and flow rate control, facilitating faster convergence in complex pipeline environments.

\subsection{Action-Decoupled SMDP-PPO}
\begin{figure}[pos=htbp]
    \centering
    \includegraphics[width=0.9\textwidth]{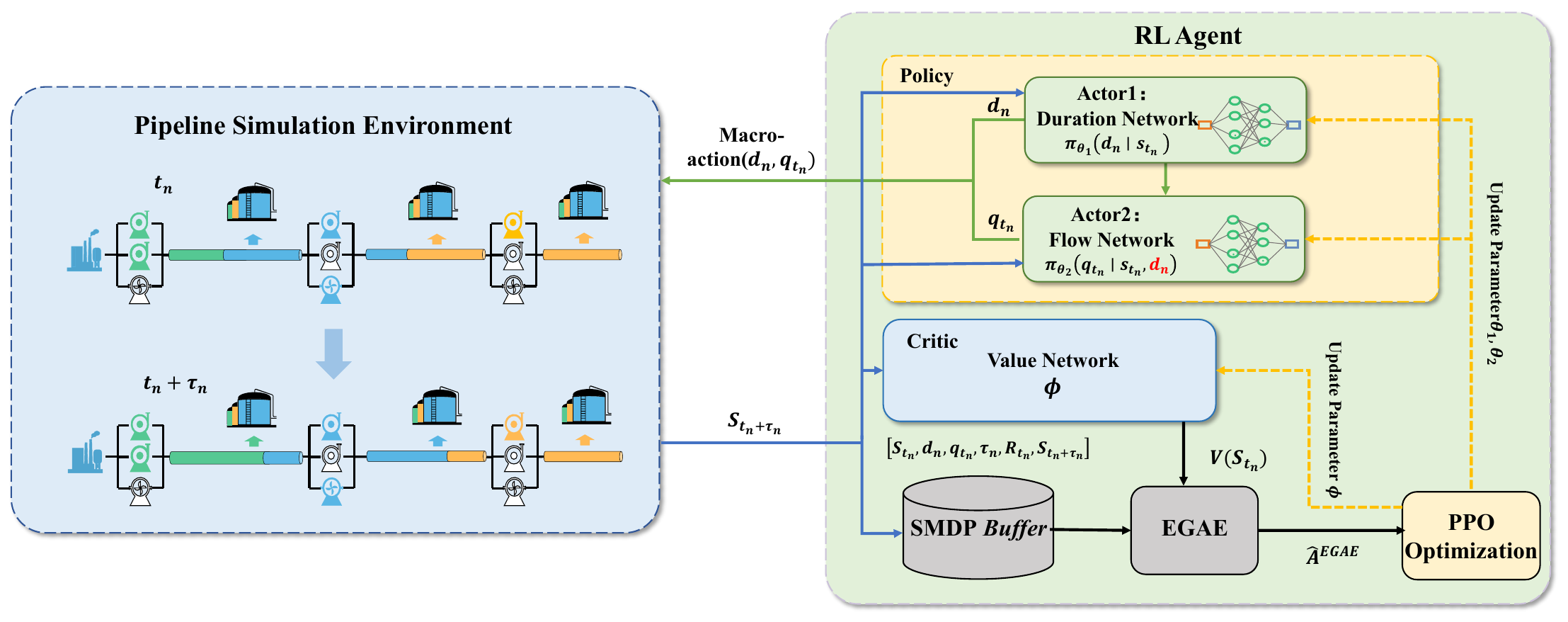}
    \caption{Interaction between the cascaded dual-actor agent and the multiproduct pipeline simulation environment under the SMDP-PPO framework.}
    \label{fig:architecture}
\end{figure}

In this subsection, the PPO algorithm is employed to obtain the optimized detailed scheduling policy within the SMDP framework. The training process adopts an actor-critic architecture, where the interaction between the dual-actor agent and the multiproduct pipeline simulation environment is illustrated in Fig. \ref{fig:architecture}. The simulation environment inherently satisfies most mathematical constraints, such as batch tracking. For constraints not directly embedded in the simulator, the delivery flow rate bounds are strictly satisfied via an action mapping mechanism that ensures all outputs are inherently feasible, while hydraulic pressure constraints are enforced via penalty terms in the reward function.

To consider the impact of execution duration in the SMDP-based scheduling model, the advantage function is estimated using extended generalized advantage estimation (EGAE)~\citep{chen2023reinforcement}. This approach accounts for the actual execution duration $\tau_n$ of each macro-action, which has been shown to balance the bias-variance trade-off in SMDP environments effectively. The temporal difference (TD) error $\delta_{t_n}$ for a scheduling transition starting at decision stage $n$ is expressed as:
\begin{equation}
\delta_{t_n} = R_{t_n} + \gamma^{\tau_{n}} V_\phi(s_{t_n+{\tau_{n}}}) - V_\phi(s_{t_n}).
\end{equation}
Based on these TD errors, the EGAE for the $n$-th decision stage is computed as:
\begin{equation}
\hat{A}_{t_n}^{E G A E}=\sum_{j=0}^{\infty} \gamma^{\sum_{l=0}^{j-1} \tau_{n+l}} \lambda^j \delta_{t_{n+j}},
\end{equation}
where $\lambda \in [0,1]$ is a hyperparameter balancing the estimation bias and variance.
The importance sampling ratio $r_{t_n}(\theta_1, \theta_2)$ for the dual-actor network is defined as:
\begin{equation}
r_{t_n}(\theta_1, \theta_2) = \frac{\pi_{\theta_1}(d_n \mid s_{t_n}) \cdot \pi_{\theta_2}(q_{t_n} \mid s_{t_n}, d_n)}{\pi_{\theta_{1}^{old}}(d_n \mid s_{t_n}) \cdot \pi_{\theta_{2}^{old}}(q_{t_n} \mid s_{t_n}, d_n)}.
\end{equation}
The parameters $\theta_1$ and $\theta_2$ are then optimized by maximizing the objective function $L^{CLIP}$:
\begin{equation}
L^{CLIP}(\theta_1, \theta_2) = \hat{\mathbb{E}}_n \left[ \min \left( r_{t_n}(\theta_1, \theta_2) \hat{A}_{t_n}^{EGAE}, \text{clip}(r_{t_n}(\theta_1, \theta_2), 1-\epsilon, 1+\epsilon) \hat{A}_{t_n}^{EGAE} \right) \right],
\end{equation}
where $\epsilon$ is the clipping coefficient. The critic network parameters $\phi$ are optimized by minimizing the value function loss:
\begin{equation}
L^{VF}(\phi) = \hat{\mathbb{E}}_n \left[ (V_\phi(s_{t_n}) - \hat{G}_{t_n})^2 \right],
\end{equation}
where $\hat{G}_{t_n}$ is the target return for stage $n$.

The detailed optimization procedure is given in Algorithm \ref{alg:smdp_ppo}. The agent iteratively samples a planned duration $d_n$ and flow rates $q_{t_n}$ through the cascaded Actor 1 and Actor 2. During the step-by-step environment execution, an event-triggered termination mechanism is employed: the simulation is interrupted if a critical event occurs (e.g., batch head arrival or pressure violation), and the actual duration $\tau_n$ is recorded. These transitions are stored in $\mathcal{D}$ for advantage estimation via EGAE followed by multi-epoch policy updates, enabling the agent to learn an optimal detailed scheduling policy under various operational constraints.

\begin{algorithm}
\caption{Action-Decoupled SMDP-PPO}
\label{alg:smdp_ppo}
\begin{algorithmic}[1]
\Require Pipeline Environment $\mathcal{E}$, Actor 1 $\pi_{\theta_1}(d | s)$, Actor 2 $\pi_{\theta_2}(q | s, d)$, Critic $V_\phi(s)$, Experience Buffer $\mathcal{D}$
\Ensure Optimized scheduling policy parameters $\theta_1^*, \theta_2^*$
\For{$\text{iteration} = 1, 2, \dots$}
    \While{$|\mathcal{D}| < \text{BATCH\_SIZE}$}
        \If{\textbf{done}}
            \State Reset environment $\mathcal{E}$, get initial state $s_{t_0}$, set $n \gets 0, t_n \gets 0$
        \EndIf
        \State \textbf{Actor 1:} Sample planned duration $d_n \sim \pi_{\theta_1}(\cdot \mid s_{t_n})$
        \State \textbf{Actor 2:} Sample flow rates $q_{t_n} \sim \pi_{\theta_2}(\cdot \mid s_{t_n}, d_n)$
        \State $R_{t_n} \gets 0, \; \tau_n \gets 0, \; R^{accum}_{int}\gets 0$
        \For{$m = 0$ \textbf{to} $d_n - 1$} \Comment{Step-by-step simulation}
            \State Step environment $\mathcal{E}$ with $q_{t_n}$, receive interval-based deviation reward $R_{int}(\cdot)$
            \State $R^{accum}_{int} \gets R^{accum}_{int} + \gamma^m R_{int}(\cdot)$ 
            \State $\tau_n \gets m + 1$
            \State Update environment termination flag \textbf{done}
            \If{\textbf{done} \textbf{or} critical event triggered} \Comment{e.g., batch head arrival or pressure violation}
                \State \textbf{break}
            \EndIf
        \EndFor
        \State Compute total stage reward: $R_{t_n} \gets \alpha R_{de} + \beta R_{sta} + \eta R_{p}$
        \State Store $(s_{t_n}, d_n, q_{t_n}, \tau_n, R_{t_n}, s_{t_n+\tau_n})$ in $\mathcal{D}$
        \State $t_{n+1} \gets t_n + \tau_n, \; n \gets n + 1$
    \EndWhile
    \State Compute advantages $\hat{A}_{t_n}^{EGAE}$ and returns $\hat{G}_{t_n}$ for all $n \in \mathcal{D}$
    \For{$\text{epoch} = 1$ \textbf{to} $E$}
        \For{mini-batch $B \subset \mathcal{D}$}
            \State Update $\theta_1, \theta_2$ by maximizing $L^{CLIP}$
            \State Update $\phi$ by minimizing $L^{VF}$
        \EndFor
    \EndFor
    \State Clear Buffer $\mathcal{D}$
\EndFor
\end{algorithmic}
\end{algorithm}

\section{Results and discussion}\label{sec5}
In this section, we conduct a case study to validate the effectiveness and efficiency of the proposed Action-Decoupled SMDP-PPO framework on a multiproduct pipeline. Furthermore, we present a dedicated comparison demonstrating the training benefits of the cascaded dual-actor architecture over a single-actor baseline. The Python 3.9-based RL models, the C\# pipeline simulator, and the Gurobi-based MILP baseline were all executed on an Intel Core i5-12450H CPU @ 2.00 GHz.

\subsection{Basic pipeline data}
A 375.5 km real-world multiproduct pipeline in China is considered for the case study, equipped with five stations: the initial station (IS), one delivery-pumping station (DPS1), two delivery stations (DS2 and DS3), and the terminal station (TS). The physical and operational parameters of these stations and the four connecting segments are listed in Tables \ref{tab:station_data_v3} and \ref{tab:segment_data}, respectively. This pipeline sequentially transports 92\# gasoline, 95\# gasoline, and 0\# diesel, whose physical properties are detailed in Table \ref{tab:product_data}. During routine operations, the pipeline is typically filled with 0\# diesel at both the beginning and the end of the scheduling horizon, and the batch injections generally follow a cyclic sequence of 92\#G-95\#G-92\#G-0\#D. The injection plan and the corresponding supply-demand requirements at each delivery station, adopted from the case study of \citet{zhou2019milpdetailed} and converted into volumetric units, are summarized in Table~\ref{tab:case1_plan}.

\begin{table}[htbp]
  \centering
  \caption{Physical and operational parameters of each station.}
  \label{tab:station_data_v3}
  \begin{tabular}{lcccc}
    \toprule
    \textbf{Station ID} 
      & \makecell{\textbf{Altitude}\\(m)} 
      & \makecell{\textbf{Min delivery}\\\textbf{flow rate} (m$^3$/h)} 
      & \makecell{\textbf{Max delivery}\\\textbf{flow rate} (m$^3$/h)} 
      & \makecell{\textbf{Max head of}\\\textbf{pump station} (m)} \\
    \midrule
    IS   & 0   & --  & --  & 1092 \\
    DPS1 & 7   & 100 & 900 & 1096 \\
    DS2  & 111 & 100 & 600 & --   \\
    DS3  & 53  & 100 & 600 & --   \\
    TS   & 51  & 100 & 600 & --   \\
    \bottomrule
  \end{tabular}
\end{table}

\begin{table}[htbp]
  \centering
  \caption{Physical and operational parameters of pipeline segments.}
  \label{tab:segment_data}
  \begin{tabular}{lcccc}
    \toprule
    \textbf{Pipeline segment} 
      & \makecell{\textbf{Length}\\(km)} 
      & \makecell{\textbf{Inner diameter}\\(mm)} 
      & \makecell{\textbf{Min allowable}\\\textbf{flow rate} (m$^3$/h)} 
      & \makecell{\textbf{Max allowable}\\\textbf{flow rate} (m$^3$/h)} \\
    \midrule
    IS--DPS1   & 130.4 & 492.2 & 600 & 1100 \\
    DPS1--DS2  & 139.5 & 492.2 & 500 & 1100 \\
    DS2--DS3   & 66.9  & 441.2 & 100 & 1100 \\
    DS3--TS    & 38.7  & 441.2 & 100 & 1100 \\
    \bottomrule
  \end{tabular}
\end{table}

\begin{table}[htbp]
  \centering
  \caption{Physical properties of refined products.}
  \label{tab:product_data}
  \begin{tabular}{ccc}
    \toprule
    \textbf{Product Type} & \textbf{Density} ($\mathrm{kg/m^3}$) & \textbf{Viscosity} ($\mathrm{m^2/s}$) \\
    \midrule
    0\# diesel       & 845 & $6.11 \times 10^{-6}$ \\
    92\# gasoline     & 740 & $2.05 \times 10^{-6}$ \\
    95\# gasoline       & 750 & $1.07 \times 10^{-6}$ \\
    \bottomrule
  \end{tabular}
\end{table}

\begin{table}[htbp]
  \centering
  \caption{Planned supply and demand for the case study.}
  \label{tab:case1_plan}
  \begin{threeparttable}
  \begin{tabular}{lcccc}
    \toprule
    \textbf{Station} & \textbf{0\# Diesel} & \textbf{92\# Gasoline} & \textbf{95\# Gasoline} & \textbf{Total} \\
     & (m$^3$) & (m$^3$) & (m$^3$) & (m$^3$) \\
    \midrule
    DPS1 & 22{,}485 & 35{,}810 & 14{,}666 & 72{,}961 \\
    DS2  &  8{,}876 & 20{,}270 &  8{,}000 & 37{,}146 \\
    DS3  & 10{,}651 & 18{,}919 &  4{,}667 & 34{,}236 \\
    TS   & 31{,}361 & 23{,}648 &  6{,}000 & 61{,}009 \\
    \midrule
    \textbf{Total demand}    & 73{,}373 & 98{,}647 & 33{,}333 & 205{,}353 \\
    \textbf{Total injection}\textsuperscript{*} & 72{,}189 & 106{,}081 & 33{,}333 & 211{,}603 \\
    \bottomrule
  \end{tabular}
  \begin{tablenotes}
    \footnotesize
    \item[*] The injection sequence consists of four batches: 92\# (52{,}027~m$^3$), 95\# (33{,}333~m$^3$), 92\# (54{,}054~m$^3$), and 0\# (72{,}189~m$^3$), while the pipeline is initially filled with 0\# diesel.
  \end{tablenotes}
  \end{threeparttable}
\end{table}

\subsection{Case study}
\subsubsection{Experimental settings}
The hyperparameters of the proposed Action-Decoupled SMDP-PPO agent are summarized in Table \ref{tab:rl_para}, with parameters for all evaluated methods carefully fine-tuned to be as optimal as possible. To benchmark the proposed framework, two baselines are employed: a standard Single-Actor SMDP-PPO model with the same hyperparameter configuration as the proposed action-decoupled framework, and a solver-based MILP baseline, obtained by reformulating the proposed MINLP model into an MILP via the piecewise linearization technique of \citet{liao2018milp} and solving it with Gurobi under the objective in Eq.~\eqref{eq:objective}. Furthermore, the choice of the discrete time window length primarily balances decision granularity against computational complexity. A large interval restricts the feasible solution space, whereas a small interval leads to a combinatorial explosion in the mathematical formulation. Consequently, the base discrete time interval for both the mathematical model and the RL environment is uniformly set to 1 hour.

\begin{table}[htbp]
  \centering
  \caption{Hyper-parameters for the Action-Decoupled SMDP-PPO Framework}
  \label{tab:rl_para}

    \begin{tabular}{lcc}
      \toprule
      \textbf{Category} & \textbf{Parameters} & \textbf{Value} \\
      \midrule
      \multirow{2}{*}{Network Architecture} 
      & Actor 1\&2 network (Layers $\times$ Neurons) & $2 \times 256$ \\
      & Critic network (Layers $\times$ Neurons) & $2 \times 256$ \\
      \midrule
      \multirow{4}{*}{PPO Algorithm} 
      & Learning rate  & $3 \times 10^{-4}$ \\
      & Discount factor ($\gamma$) & 0.99 \\
      & PPO clip coefficient ($\epsilon$) & 0.2 \\
      & GAE parameter ($\lambda$) & 0.95 \\
      \midrule
      \multirow{1}{*}{SMDP Setting} 
      & Maximum decision duration ($d^{\max}$) & 30 \\
      \midrule
      \multirow{6}{*}{Reward Weights} 
      & Delivery deviation reward weight ($\alpha$) & 1.0 \\
      & Flow rate stability weight ($\beta$) & 0.5 \\
      & Pressure penalty weight ($\eta$) & 5 \\
      & NRMSE threshold ($\delta$) & 0.15 \\
      & Terminal penalty coefficient ($\Gamma$) & 500 \\
      & Switching penalty ($C_{switch}$) & 2 \\
      
      \bottomrule
    \end{tabular}

\end{table}

\subsubsection{Performance comparison and scheduling results}
\begin{figure}[pos=htbp]

    \centering
    \includegraphics[width=0.9\textwidth]{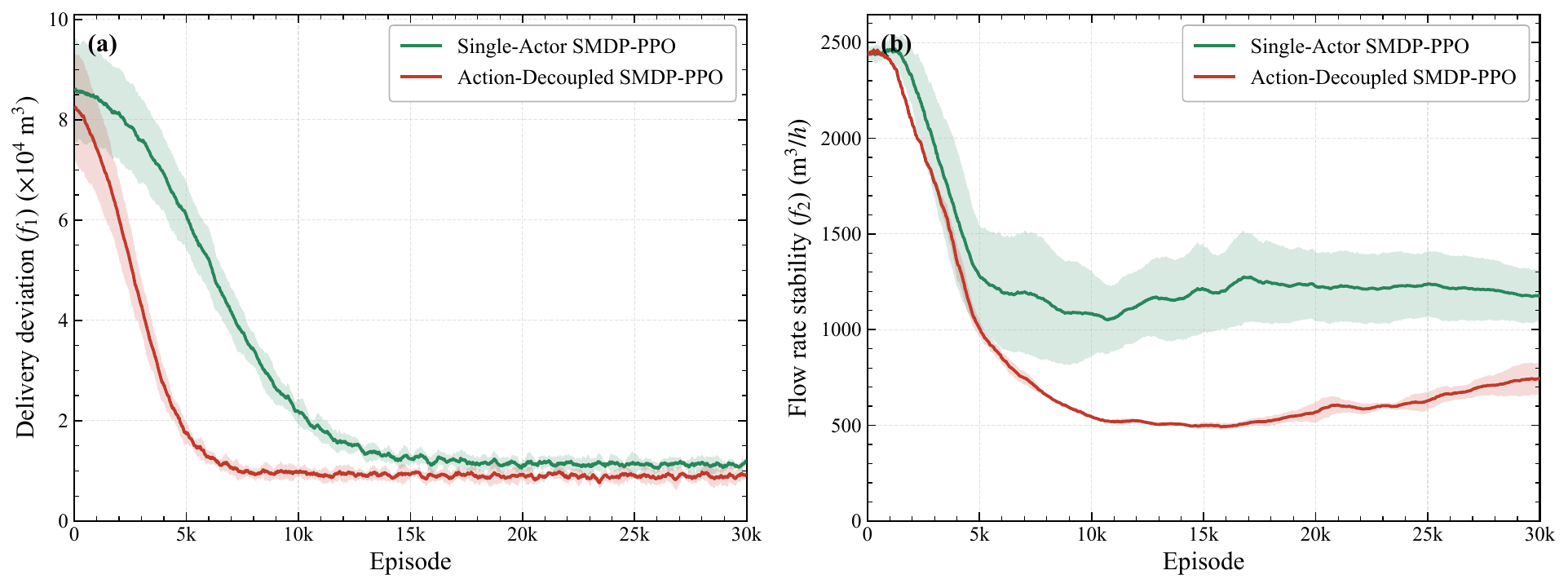}
    \caption{Comparison of training performance between the proposed Action-Decoupled and Single-Actor SMDP-PPO. (a) Convergence of the delivery deviation $f_1$; (b) convergence of the flow rate stability $f_2$. The solid line and shaded band denote the mean and standard deviation over three independent runs.}
    \label{fig:convergence}
\end{figure}

Fig.~\ref{fig:convergence} presents the training curves of the proposed Action-Decoupled SMDP-PPO and the Single-Actor SMDP-PPO baseline with respect to the two optimization objectives, i.e., the delivery deviation $f_1$ (Fig.~\ref{fig:convergence}(a)) and the flow rate stability $f_2$ (Fig.~\ref{fig:convergence}(b)). Specifically, the total planned injection exceeds the total demand by 6{,}250~m{$^3$}(about 3.0\% of the total demand) in this case, which makes a zero delivery deviation unattainable and sets the lower bound of $f_1$ for this case. Despite this, Fig.~\ref{fig:convergence}(a) shows that both methods stabilize the total delivery deviation to below 5\% of the total demand during the mid-to-late training stage. A schedule reaching this 5\% threshold is regarded as feasible in this work, since the surplus injection makes a lower deviation hardly attainable. Both methods reach this level, confirming that the SMDP formulation effectively optimizes the detailed scheduling policy. Furthermore, the proposed method begins to stabilize around 5{,}000 episodes, markedly earlier than the single-actor baseline. A consistent advantage is observed for $f_2$ in Fig.~\ref{fig:convergence}(b): the proposed method converges to a substantially lower flow rate fluctuation than the baseline, indicating smoother segment flow rates and thus a more practical schedule. This contrast verifies that decoupling the discrete duration and continuous flow rate via two cascaded actors eliminates the gradient interference inherent in the single-actor scheme, thereby improving both convergence speed and final accuracy. A more comprehensive comparison between the two architectures is provided in Section~\ref{secablation}.

\begin{figure}[pos=htbp]
    \centering
    \includegraphics[width=0.9\textwidth]{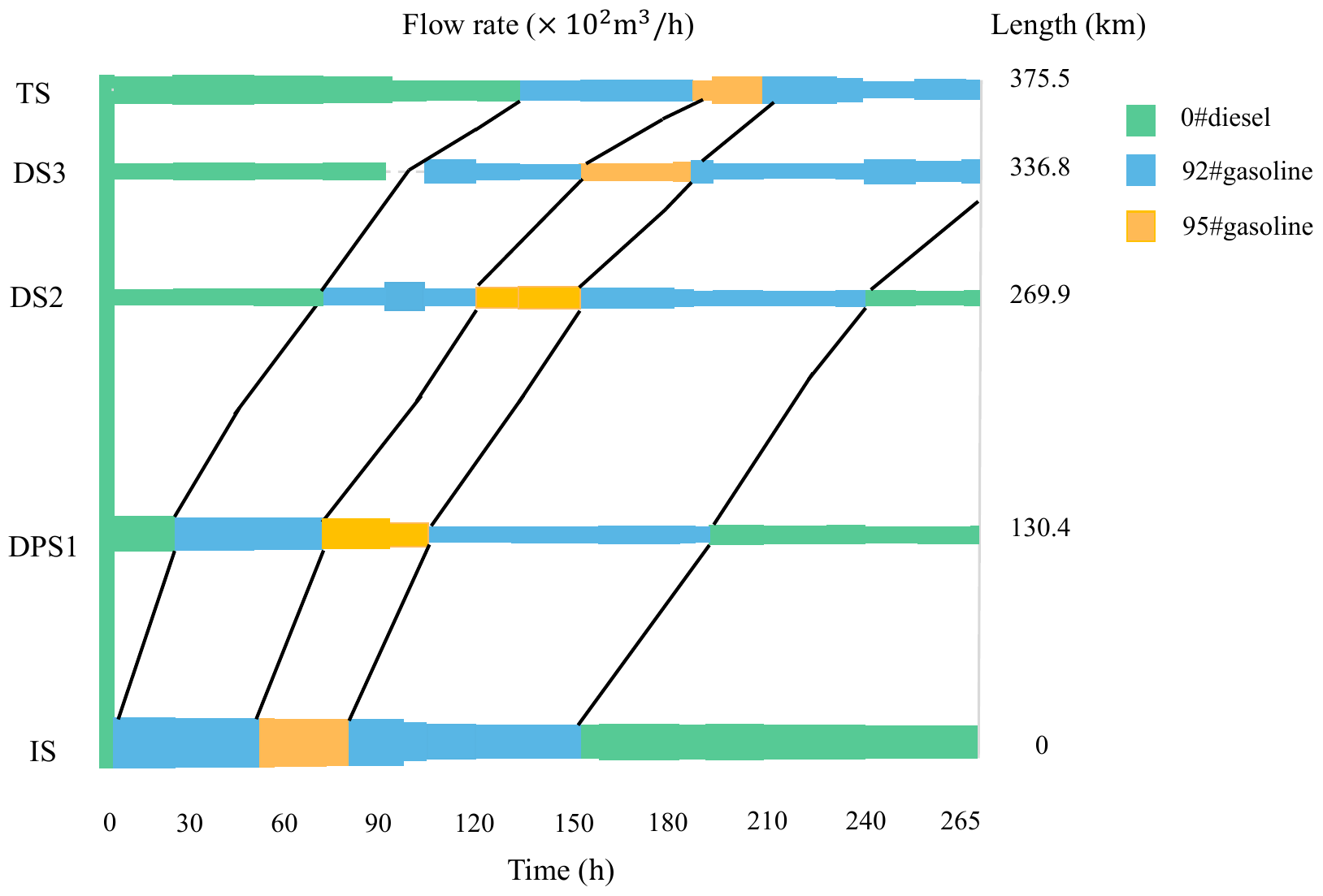}
    \caption{Batch migration chart of a representative detailed scheduling scheme found by the Action-Decoupled SMDP-PPO during training.}
    \label{fig:gantt}
\end{figure}

\begin{figure}[pos=htbp]
    \centering
    \includegraphics[width=0.9\textwidth]{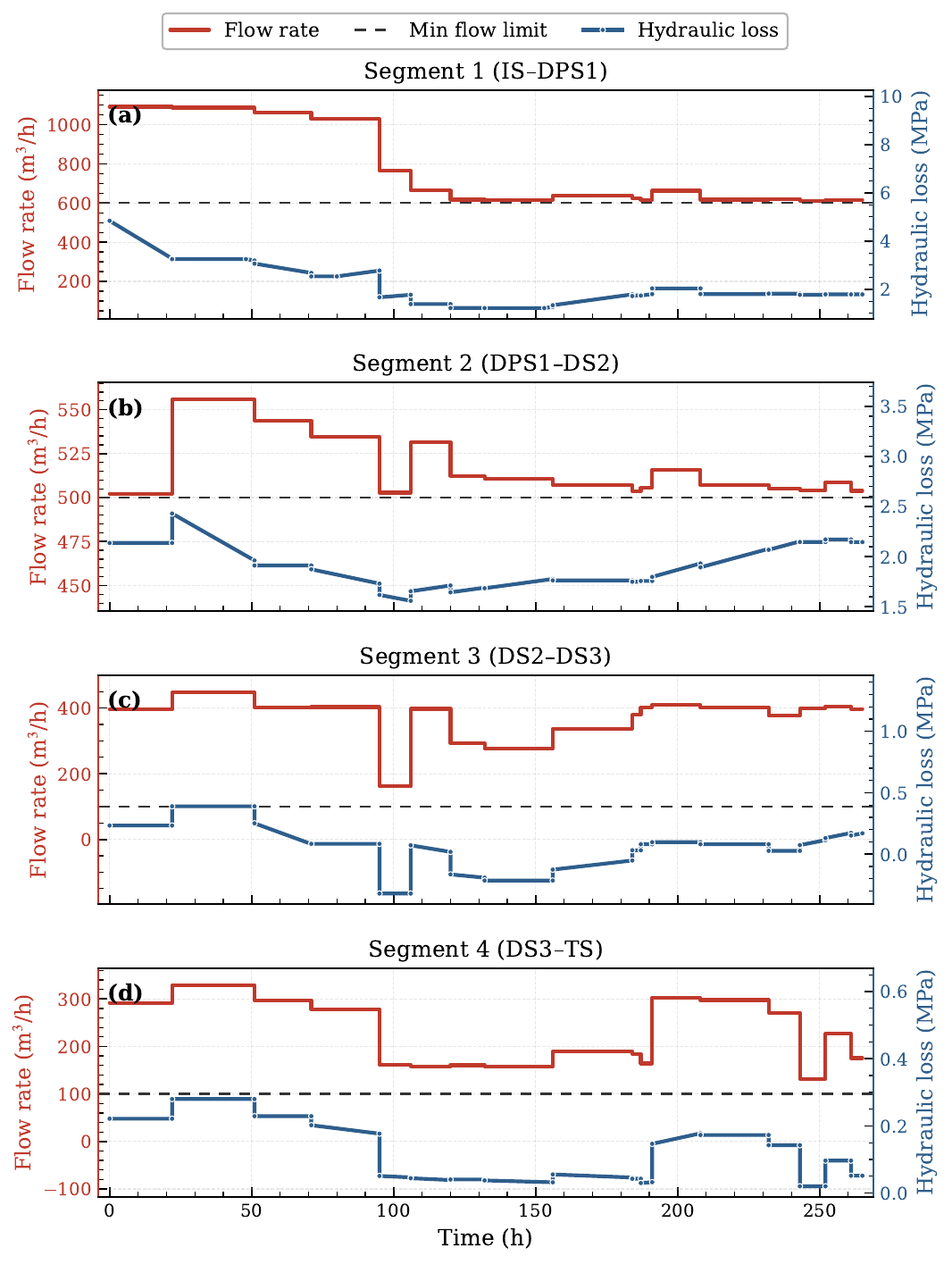}
    \caption{Flow rate and hydraulic loss of each pipeline segment under the obtained scheduling scheme. The red step lines and blue lines denote the segment flow rate and hydraulic loss, respectively, and the dashed lines indicate the minimum flow rate limits. (a) Segment 1 (IS–DPS1); (b) Segment 2 (DPS1–DS2); (c) Segment 3 (DS2–DS3); (d) Segment 4 (DS3–TS).}
    \label{fig:flow_hydraulic_loss}
\end{figure}

Fig.~\ref{fig:gantt} depicts a representative detailed scheduling scheme found by the Action-Decoupled SMDP-PPO, presented as a batch migration chart. In this chart, the vertical and horizontal axes correspond to the length coordinate of each station and the operating time, respectively. Each black line traces the trajectory of a batch interface along the pipeline, with its slope indicating the prevailing pipeline flow rate. The injection and delivery operations are marked by colored blocks, where the color identifies the product type and the block thickness scales with the corresponding flow rate. The segment flow rates and hydraulic losses corresponding to this scheme are further reported in Fig.~\ref{fig:flow_hydraulic_loss}. Owing to the sequential action mapping mechanism embedded in the output layer of Actor~2, the flow rate of every pipeline segment is kept strictly within its allowable bounds over the entire horizon. Meanwhile, as the batch interfaces migrate, the hydraulic loss of each segment varies with the changing product distribution. Specifically, the hydraulic loss of Segment~3 (DS2--DS3) turns negative during certain low-flow intervals, where its negative elevation-related pressure drop $P^H$ (a gravitational head gain from the 58$\mathrm{~m}$ descent) outweighs the reduced frictional pressure drop $P^F$. At the system level, the cumulative pressure demand along the pipeline stays below the supply capacity of the upstream pump stations, thereby satisfying the pressure constraints in Eqs.~\eqref{eq:p_limit1} and \eqref{eq:p_limit2}. These results confirm that the proposed method yields an operationally feasible schedule that satisfies both the flow rate and pressure constraints. To further evaluate the quality of this feasible schedule, its scheduling objectives and computational cost are compared against both baselines.

Table~\ref{tab:delivery_result} reports the actual delivered volumes at each station, and Table~\ref{tab:method_comparison} summarizes the overall performance of the three methods, where the schedules of the two RL-based methods correspond to representative results that are attained by the trained policies. Since no cross-plan generalization is considered, each plan is solved by training a dedicated policy from scratch, and the reported solution time accordingly refers to the training time required to reach that schedule. In terms of delivery accuracy, the Gurobi-based MILP model attains the lowest deviation $f_1 = 6,250~\mathrm{m}^3$, exactly reaching the inherent lower bound set by the surplus injection, which amounts to about 3.0\% of the total demand. In contrast, the proposed method reaches 7{,}153$\mathrm{~m}^3$, about 3.5\% of the total demand, closely approaching this bound and clearly outperforming the single-actor baseline at 8,442$\mathrm{~m}^3$. For the flow rate stability $f_2$, however, the proposed method gives the smallest value of 561$\mathrm{~m^3/h}$, markedly better than the single-actor baseline at 1,449$\mathrm{~m^3/h}$ and the Gurobi-based model at 1,697$\mathrm{~m^3/h}$. Notably, although Gurobi drives $f_1$ to its lower bound, it is terminated with a 13.8\% optimality gap after 38{,}911$\mathrm{~s}$, and since the deviation term is already optimal, this residual gap is essentially attributable to the unoptimized flow rate stability, which explains its poor $f_2$ and would require substantially longer computation to be further reduced. The advantage of the proposed method lies in its computational efficiency: it produces a feasible schedule in about 900$\mathrm{~s}$ and a representative best schedule in about 1,503$\mathrm{~s}$. The full training of 30{,}000 episodes takes about 3{,}000~s in total, beyond which no further improvement is observed. Overall, the proposed method achieves delivery accuracy close to the exact optimum, along with the stable flow rate, while reducing solution time by more than an order of magnitude, making it well suited for the efficient detailed scheduling of multiproduct pipelines.

\begin{table*}[htbp]
  \centering
  \caption{Planned demand and actual delivered volumes at each station for the representative schedules found by the three methods during training/solving.}
  \label{tab:delivery_result}
  \small
  \setlength{\tabcolsep}{4pt}
  \begin{threeparttable}
  \begin{tabular}{l *{3}{cccc}}
    \toprule
    \multirow{3}{*}{\textbf{Station}}
      & \multicolumn{4}{c}{\textbf{0\# Diesel}}
      & \multicolumn{4}{c}{\textbf{92\# Gasoline}}
      & \multicolumn{4}{c}{\textbf{95\# Gasoline}} \\
      & \multicolumn{4}{c}{(m$^3$)}
      & \multicolumn{4}{c}{(m$^3$)}
      & \multicolumn{4}{c}{(m$^3$)} \\
    \cmidrule(lr){2-5}\cmidrule(lr){6-9}\cmidrule(lr){10-13}
      & Demand & Prop. & SA & Gurobi
      & Demand & Prop. & SA & Gurobi
      & Demand & Prop. & SA & Gurobi \\
    \midrule
    DPS1 & 22{,}485 & 22{,}875 & 22{,}625 & 22{,}485
         & 35{,}810 & 35{,}829 & 35{,}045 & 35{,}810
         & 14{,}666 & 14{,}719 & 14{,}651 & 14{,}666 \\
    DS2  &  8{,}876 & 10{,}674 & 11{,}861 & 15{,}005
         & 20{,}270 & 20{,}219 & 20{,}333 & 20{,}391
         &  8{,}000 &  8{,}280 &  7{,}694 &  8{,}000 \\
    DS3  & 10{,}651 & 11{,}082 & 11{,}434 & 10{,}651
         & 18{,}919 & 19{,}088 & 19{,}678 & 18{,}919
         &  4{,}667 &  4{,}717 &  4{,}803 &  4{,}667\\
    TS   & 31{,}361 & 35{,}031 & 33{,}407 & 31{,}361
         & 23{,}648 & 23{,}681 & 23{,}998 & 23{,}648
         &  6{,}000 &  5{,}792 &  6{,}093 &  6{,}000 \\
    \midrule
    \textbf{Abs. dev.}
      & -- & 6{,}289 & 5{,}954 & 6{,}129
      & -- &   272 & 1{,}937 &   121
      & -- &   592 &   551 &     0 \\
    \bottomrule
  \end{tabular}
  \begin{tablenotes}
    \footnotesize
    \item Prop. = Action-Decoupled SMDP-PPO; SA = Single-Actor SMDP-PPO. ``Abs. dev.'' is the absolute delivery deviation summed over all stations for each product. 
  \end{tablenotes}
  \end{threeparttable}
\end{table*}

\begin{table}[htbp]
  \centering
  \caption{Performance comparison of the proposed method, the single-actor baseline, and the Gurobi-based MILP model.}
  \label{tab:method_comparison}
  \begin{tabular}{lcccc}
    \toprule
    \textbf{Method} & $\boldsymbol{f_1}$ ($\mathrm{m^3}$) & $\boldsymbol{f_2}$ ($\mathrm{m^3/h}$) & \textbf{Solution time} ($\mathrm{s}$) & \textbf{Gap} ($\mathrm{\%}$) \\
    \midrule
    Action-Decoupled SMDP-PPO & 7{,}153 & 561   & 1{,}503 & -- \\
    Single-Actor SMDP-PPO     & 8{,}442 & 1{,}449 & 2{,}059 & -- \\
    Gurobi                    & 6{,}250 & 1{,}697 & 38{,}911      & 13.8 \\
    \bottomrule
  \end{tabular}
\end{table}

\subsubsection{Effect of action decoupling}\label{secablation}
\begin{figure}[pos=htbp]
    \centering
    \includegraphics[width=0.9\textwidth]{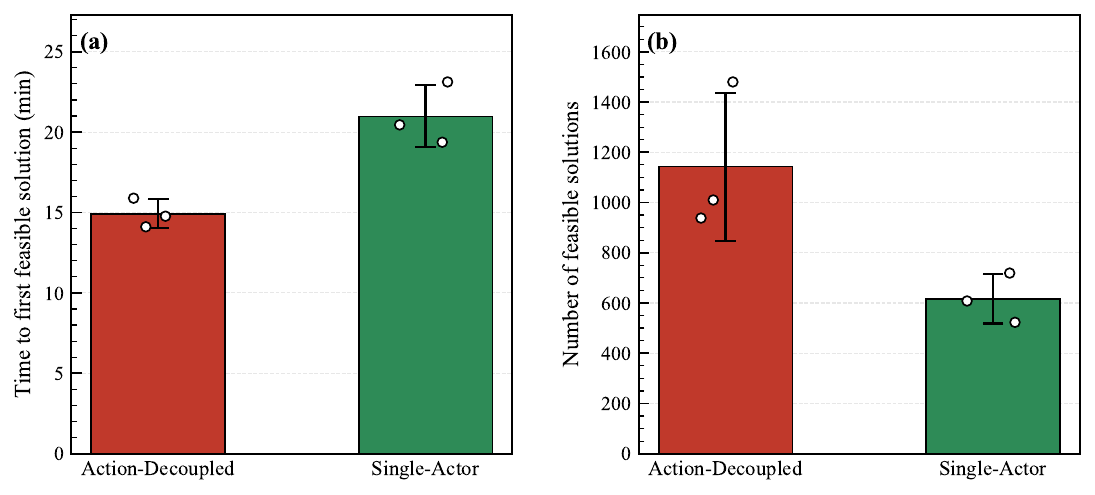}
    \caption{Comparison between the Action-Decoupled and Single-Actor SMDP-PPO. (a) Time to the first feasible solution; (b) number of feasible solutions over 30{,}000 episodes. Bars and error bars denote mean and standard deviation over three runs; circles mark individual runs.}
    \label{fig:ablation}
\end{figure}

To further quantify the benefit of the action-decoupled architecture beyond the convergence trends in Fig.~\ref{fig:convergence}, two training-process metrics are compared between the proposed method and the single-actor baseline over three independent runs, as shown in Fig.~\ref{fig:ablation}. The first metric is the time to reach the first feasible solution, which measures training efficiency. As shown in Fig.~\ref{fig:ablation}(a), the proposed method obtains a feasible schedule in about 14.9$\mathrm{~min}$ on average, markedly faster than the single-actor baseline at about 21.0$\mathrm{~min}$, and with a smaller variation across runs. The second metric is the number of feasible solutions found during the 30{,}000-episode training, which reflects the stability of the policy in maintaining feasibility once it is reached. As shown in Fig.~\ref{fig:ablation}(b), the proposed method produces substantially more feasible solutions in every run, with an average of 1{,}143 compared with 617 for the baseline, indicating that, once a feasible region is found, the decoupled policy remains within it far more consistently. These results confirm that decoupling the discrete duration and the continuous flow rate into two cascaded actors not only accelerates convergence but also improves the reliability of the learned policy.

\section{Conclusion}\label{sec6}
This paper proposed an Action-Decoupled SMDP-PPO framework that efficiently solves the detailed scheduling of multiproduct pipelines. The scheduling problem was first formulated as a hybrid-time MINLP model that jointly minimizes the delivery deviation and the flow rate fluctuation under rigorous batch tracking and hydraulic constraints, and was then translated to a semi-Markov decision process so that reinforcement learning could be applied directly. On a real-world pipeline case in China, the learned policy kept the delivery deviation close to the lower bound reached by the linearized MILP model while reducing the solution time by more than an order of magnitude. A cascaded dual-actor architecture was further designed to decouple the discrete operating duration from the continuous flow rates, removing the gradient interference that destabilizes conventional single-actor policies; this decoupling yielded a markedly lower flow rate fluctuation and, as confirmed by our experiments, accelerated convergence and improved the reliability of the learned policy.

Despite these results, the current study trains a dedicated policy for each scheduling plan without cross-plan generalization, on a single-source pipeline topology. Future work will develop generalizable policies that adapt to varying demand plans without retraining, and extend the framework to tree-structured and multi-source pipeline networks to further enhance its practical applicability.

\section*{Acknowledgments}
This work was supported by the National Key Research and Development Program of China under Grant Nos. 2021YFA1000104 and 2021YFA1000100, the Laboratory Project of Higher Education Institutions in Shandong Province---Energy System Intelligent Management and Policy Simulation Laboratory at China University of Petroleum, the Youth Innovation Team of Higher Education Institutions in Shandong Province---Data Intelligence Innovation Team at China University of Petroleum, and the Innovation Fund Project for Graduate Students of China University of Petroleum (East China) through the Fundamental Research Funds for the Central Universities under Grant No. 25CX04042A.

\bibliographystyle{cas-model2-names}
\bibliography{cas-refs}

\appendix
\section{Nomenclature}\label{app:nomenclature}
\begin{tcolorbox}[colback=white,colframe=black,breakable]
\printnomenclature
\nomenclature[A]{$k \in K$}{Index set of time nodes, $k = 1, 2, \dots, k_{\max}$}
\nomenclature[A]{$t_k \in T$}{Time value of node $k$ ($h$)}

\nomenclature[A]{$i \in I$}{Set of stations and pipeline segments. Pipeline segment $i$ connects station $i$ with station $i+1$, $i=1,2, \cdots, i_{\max}$}
\nomenclature[A]{$j \in J$}{Set of batches, $j=1,2, \cdots, j_{\max}$}
\nomenclature[A]{$J^{old}$}{The set of batches in the pipeline at the start of the scheduling horizon}
\nomenclature[A]{$J^{new}$}{The set of batches to be injected into the pipeline during the scheduling horizon}
\nomenclature[A]{$o \in O$}{Set of product types}
\nomenclature[A]{$J_o$}{The set of batches that belong to product type $o$}

\nomenclature[B]{$V^x_{i, o}$}{Demand volumes for product $o$ at delivery station $i$ $\left(\mathrm{m}^3\right)$}
\nomenclature[B]{$V^d_{i, o}$}{Actual delivered volumes during the whole study horizon for product $o$ at delivery station $i$}
\nomenclature[B]{$\lambda_{i}$}{Weight factor of pipeline segment $i$ used to prevent flow rate fluctuations from being concentrated in specific segments}
\nomenclature[B]{$Q^p_{k, i}$}{Flow rate of pipeline segment $i$ during time interval $\left(t_k, t_{k+1}\right)$ $\left(\mathrm{m}^3 / \mathrm{h}\right)$}
\nomenclature[B]{$V^s_{i}$}{the volumetric coordinate of the delivery station $i$ $\left(\mathrm{m}^3\right)$}
\nomenclature[B]{$V_{k, j}^b$}{The volumetric coordinate of the head interface of batch $j$ at time node $k$ $\left(\mathrm{m}^3\right)$}
\nomenclature[B]{$Q_k^{in}$}{The injection flow rate at the initial station during time interval  $\left(t_k, t_{k+1}\right)$ $\left(\mathrm{m}^3 / \mathrm{h}\right)$}
\nomenclature[B]{$Q_{k, i}^d$}{The delivery flow rate at station $i$ during time interval $\left(t_k, t_{k+1}\right)$ $\left(\mathrm{m}^3 / \mathrm{h}\right)$}
\nomenclature[B]{$Q^{d,\min}_i, Q^{d,\max}_i$}{Minimum and maximum allowable delivery flow rates at station $i$ $\left(\mathrm{m}^3 / \mathrm{h}\right)$}
\nomenclature[B]{$Q^{p,\min}_i, Q^{p,\max}_i$}{Minimum and maximum allowable flow rates for pipeline segment $i$ $\left(\mathrm{m}^3 / \mathrm{h}\right)$}
\nomenclature[B]{$\rho_j$}{Density of batch $j \left(\mathrm{kg}/\mathrm{m}^3\right)$}
\nomenclature[B]{$L_{k,i,j}$}{Physical length of batch $j$ in segment $i$ at time node $k$ $(\mathrm{m})$}
\nomenclature[B]{$\nu_j$}{Kinematic viscosity of batch $j\left(\mathrm{m}^2/\mathrm{s}\right)$}
\nomenclature[B]{$d_i$}{Inner diameter of pipeline segment $i$ $(\mathrm{m})$}
\nomenclature[B]{$H_{k,i,j}$}{Elevation difference spanned by batch $j$ in segment $i$ at time node $k$ $(\mathrm{m})$}
\nomenclature[B]{$P^F_{k, i}$}{Frictional pressure drop of pipeline segment $i$ at time node $k$ $(\mathrm{Pa})$}
\nomenclature[B]{$P^H_{k, i}$}{Elevation pressure drop of pipeline segment $i$ at time node $k$ $(\mathrm{Pa})$}
\nomenclature[B]{$P^{\max}_i$}{Maximum allowable pressure provided by station $i$ $(\mathrm{Pa})$}
\nomenclature[B]{$E_i$}{Elevation of station $i$ $(\mathrm{m})$}

\nomenclature[C]{$S_{k, i, j}^{head}$}{1 if the head interface of batch $j$ has passed station $i$ at time node $k$; 0 otherwise}
\nomenclature[C]{$S_{k, i, j}^{cover}$}{1 if batch $j$ covers station $i$ (available for delivery) at time node $k$; 0 otherwise }
\nomenclature[C]{$Z^{thru}_{k,i,j}$}{1 if the head interface of batch $j$ has passed through station $i+1$ but the tail interface has not passed through station $i$ at time node $k$; 0 otherwise }
\nomenclature[C]{$Z^{H\_in}_{k,i,j}$}{1 if only the head interface of batch $j$ is in pipeline segment $i$ at time node $k$; 0 otherwise}
\nomenclature[C]{$Z^{T\_in}_{k,i,j}$}{1 if only the tail interface of batch $j$ is in pipeline segment $i$ at time node $k$; 0 otherwise }
\nomenclature[C]{$Z^{all\_in}_{k,i,j}$}{1 if the whole batch $j$ is fully inside pipeline segment $i$ at time node $k$; 0 otherwise }
\nomenclature[C]{$S^{d}_{k, i}$}{1 if station $i$ delivers products during time interval $\left(t_k, t_{k+1}\right)$; 0 otherwise}
\nomenclature[C]{$B_{k}$}{1 if a head interface reaches a station at time node $k$; 0 otherwise}
\end{tcolorbox}

\end{document}